\documentclass{article}[15pt]
\usepackage{amsmath,epsfig,calc,rotating}
\usepackage{graphics}
\usepackage{wrapfig}
\usepackage{ulem}

\usepackage{natbib}
\bibpunct{(}{)}{;}{a}{}{,}

\begin{document}

\title{
Fitting Ranked English and Spanish Letter Frequency Distribution 
in U.S. and Mexican Presidential Speeches
\vspace{0.2in}
\author{
Wentian Li$^{1,*}$ and  Pedro Miramontes $^{2,3}$\\
{\small \sl 1. The Robert S. Boas Center for Genomics and Human Genetics,
The Feinstein Institute for Medical Research} \\
{\small \sl North Shore LIJ Health System,
Manhasset, 350 Community Drive, NY 11030, USA.}\\
{\small \sl 2. Departamento de Matem\'{a}ticas, Facultad de Ciencias }\\
{\small \sl Universidad Nacional Aut\'{o}noma de M\'{e}xico,
Circuito Exterior, Ciudad Universitaria,  M\'{e}xico 04510 DF, M\'{e}xico.}\\
{\small \sl 3. Centro de Ciencias de la Complejidad,
Universidad Nacional Aut\'{o}noma de M\'{e}xico} \\
{\small \sl Circuito Escolar, Ciudad Universitaria, M\'exico 04510 DF, M\'exico.}
}
\date{}
}
\maketitle  
\markboth{\sl Li and Miramontes) }{\sl Li and Miramontes}

\vspace{0.3in}

\begin{center}
{\bf ABSTRACT}: 
\end{center}

The limited range in its abscissa of ranked letter frequency distributions
causes multiple functions to fit the observed distribution reasonably well.
In order to critically compare various functions, we apply the
statistical model selections on ten functions, using the texts of U.S. and Mexican
presidential speeches in the last 1-2 centuries. Dispite minor switching
of ranking order of certain letters during the temporal evolution
for both datasets, the letter usage is generally stable. The best fitting
function, judged by either least-square-error or by AIC/BIC
model selection, is the Cocho/Beta function. We also use a novel method
to discover clusters of letters by their observed-over-expected frequency ratios.

\newpage

\large

\section{Introduction}

\indent

Although morphemes, not letters, are usually considered to be 
the smallest linguistic unit, studying statistics of letter usage 
has its own merit. For example, information on letter frequency is 
essential in cryptography for deciphering a substitution code \citep{friedman},
and ``frequency analysis" was used in as early as the 9th century 
by the Arab scientist al-Kindi for the purpose of decryption \citep{al-kindi}.

An efficient design of a communication code also depends 
crucially on the letter frequency.  The shortest Morse 
code is reserved to letters that are the most common: 
one dot for letter {\sl e} and one dash for letter {\sl t}, 
both letters being the most frequent in English.
The same principle is also behind the design of minimum-redundancy 
code by Huffman \citep{huffman}. 

The initial motivation for the ``QWERTY" mechanical typewriter design is to keep the most
common letters far away in the keyboard so that metal bars would not
jam for a fast typist \citep{david}. Even in modern times,
the digraph (letter pairs) frequency is an important piece
of information for keyboard design \citep{zhai}.

In all these examples, a quantitative description of
letter usage frequency is important. Unlike the ranked word 
frequency distribution, which is well characterized by a simple 
power-law function or Zipf's law \citep{zipf}, it is not 
clear whether a universal fitting function exists despite 
a claim of such a function (the logarithmic function) in \citep{kanter}.

In this paper, we aim at critically examining various
functional forms of fitting rank-frequency distribution
of letters, ranging from simple to more complicated
ones with two or three free parameters. The dataset used is
the historical U.S. and Mexican presidential speeches.
The presidential speeches are readily available
(see another study where the Italian presidential ``end of year"
addresses are used \citep{tuzzy}), they also offer an 
opportunity for investigations of temporal patterns in 
letter usage.

The ranked word frequency distributions studied by George Zipf
have extremely long tails, due to the presence of 
low-frequency words (such as {\sl hapax legomena}).  As a result, 
logarithmic transformation is usually applied to the $x$-axis 
(as well as the $y$-axis). The double logarithmic transformation 
is also justified by the expectation of a power-law 
function, as it will lead to a linear regression.
This linear fitting in log-log scale may have its pitfall
\citep{newman}, one being the uneven distribution of points 
along the log-transformed $x$-axis. 

For ranked letter frequency data, the finite number of 
alphabets sets an upper bound for the rank, and there 
is no large number of rare events which is an important
theoretical issue in modeling the word rank-frequency
distribution \citep{baayen}.  On the other hand, the
limited range of abscissa may make it hard to distinguish
different fitting functions. Since power-law function 
is not expected to be the best fitting function,
double logarithmic transformation is not necessary,
and we will fit the data in linear-linear scale.
No longer linear fittings, the curve fitting is
carried out  by nonlinear least-square \citep{nls}.

Statistical models with a larger number of free parameters will
guarantee to fit the data better than a model subset with fewer
number of parameters. To compare the performance of models
with different number of parameters, a penalty should be
imposed on the extra number of parameters. Towards this end,
we apply the standard model selection technique with
Akaike Information criterion or AIC \citep{aic,aic-book} and
Bayesian Information Criterion or BIC \citep{bic} to compare
various functions used to fit the ranked letter frequency data.

\section{Data}

\indent

{\bf US presidential inaugural speeches:}
In order to take into account of any possible letter usage
trend in time, we use the US Presidential Inaugural
Speech texts for the 44 presidents in the last 200 years.
The data is downloaded from the {\sl The American Presidency
Project} from the University of California at Santa Barbara
site ({\sl http://www.presidency.ucsb.edu/}).
Multiple inaugural speeches from the same person are combined
into one, including the nonconsecutive presidency of
Grover Cleveland. Five presidents did not give an inaugural
speech (John Tyler, Millard Fillmore, Andrew Johnson, Chester Arthur, Gerald Ford).
The final dataset consists of 38 text files.

{\bf Mexico presidential addresses to the congress:}
For Spanish texts, we selected the 19 Mexican presidents'
report to congress ({\sl Informes Presidenciales}) from 1914 to 2006. 
\\
({\sl http://www.diputados.gob.mx/cedia/sia/re\_info.htm} )
Again, addresses by the same president are combined into
one text file. Some presidential texts are much shorter than others
due to two possible reasons: either did the president 
only present one address (the typical number of addresses is 6), 
such as Adolfo de la Huerta (1920) and Emilio Portes Gil (1929), 
or the president gave shorter reports, such as Ernesto 
Zedillo Ponce de Le\'{o}n (1995-2000) and Vicente Fox Quesada (2001-2006).

\section{Letter frequencies and their temporal trends}

\indent

\begin{figure}[th]
\begin{center}
  \begin{turn}{-90}
     \epsfig{file=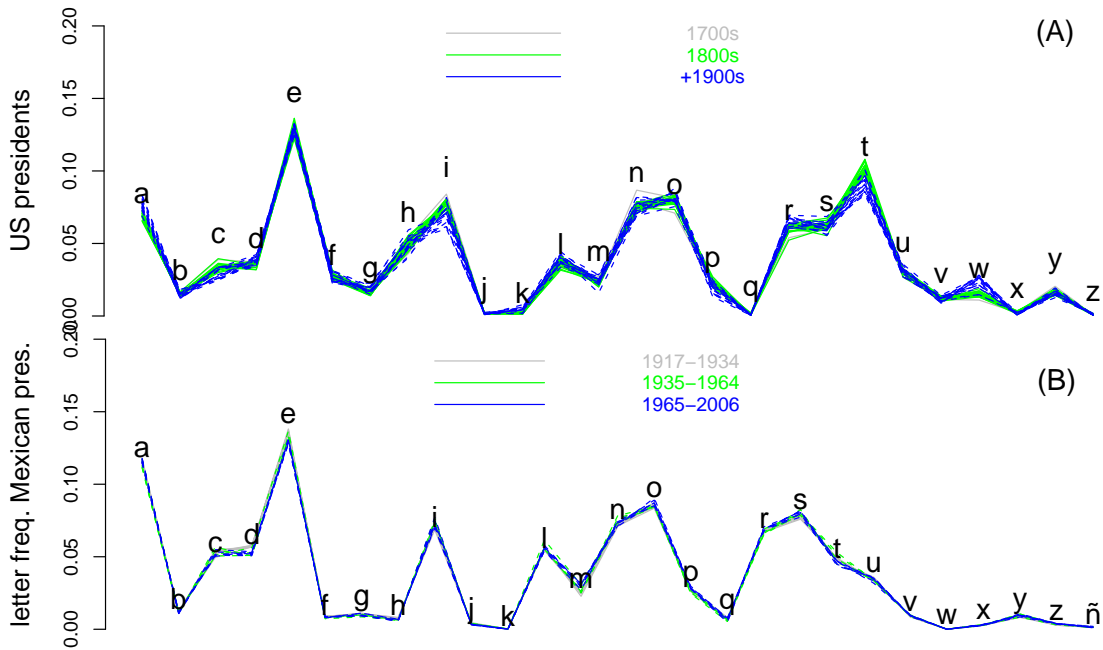, width=8.5cm}
  \end{turn}
\end{center}
\caption{
\label{fig1}
English (A) and Spanish (B) letter frequencies (unranked,
in alphabetic order) for 38 U.S. presidential inaugural
speeches and 19 Mexican presidents' report to congress.
Letter frequencies of each president's speech are linked
by a line, and different time periods are drawn separately 
(U.S. president speech: 1789-1800, shifting 1801-1901 by
0.02, shifting 1905-now by 0.04; Mexico president speech:
1919-1934, shifting 1935-1964 by 0.02, shifting 1965-2005
by 0.04). Due to a larger sample size, the fluctuation of 
frequency from president to president in Spanish texts is much smaller
than that in English texts.
}
\end{figure}

Fig.\ref{fig1}(A) shows the English letter frequency of the 38 US president's speeches,
separated by the century. The letter {\sl e} remains the most
commonly used English letter with little change in its frequency.
However there seems to be a trend of less usage of letter {\sl t},
and more usage of letter {\sl w} in the 20th century as compared to
the 19th century.

Similar frequencies of Spanish letters in the 19 Mexican presidents'
addresses are shown in Fig.\ref{fig1}(B).  Letters with accent 
(the acute accent for the vocals, the umlaut for
the letter  \"{u} and the tilde for the \~{n})
are counted separatedly but later combined because they do not 
really represent different letters.
The 19 files are arbitrarily split into three groups: 
the first 7 presidents (from 1917 to 1934), the next 5 presidents 
(from 1935 to 1964), and the last 7 presidents (from 1965 to 2006). 
These three groups are separatedly drawn in Fig.\ref{fig1}(B).
The narrowing of the variations of letter frequency in Fig.\ref{fig1}(B) 
as compared to Fig.\ref{fig1}(A) is due to the larger sample sizes 
in Spanish texts.

\begin{table}
\small
\begin{tabular}{ccccc}
\hline
 & name & year(s) of speech & sorting of alphabets & num \\
\hline
1 & Washington    & 1789,1793	& etinoashrcdlumfpybwgvxjqkz &     7710 \\
2 & Adams         & 1797	& etnioasrhdlcfumpgybvwxjkqz &     11281 \\
3 & Jefferson     & 1801,1805	& etoinasrhldcufmpwgybvkxjzq &     18701 \\
4 & Madison       & 1809,1813	& etoinasrhldcufpmgwbyvkxjzq &     11572 \\
5 & Monroe        & 1817,1821	& etinoarshdclufpmwygbvxkjzq &     37522 \\
6 & Adams         & 1825	& etoinasrhdlcfupmgybvwxjqkz &     14572 \\
7 & Jackson       & 1829,1833	& etoinarshldcufmpybgwvxkjqz &     11372 \\
8 & Van Buren     & 1837	& etoinasrhlducfpmywgvbxkjqz &     19215 \\
9 & Harrison      & 1841	& etoinarshcdlfumpygbwvxkjzq &     40526 \\
10 & Polk         & 1845	& etoniasrhdlcufmpygbwvxjqkz &     23475 \\
11 & Taylor       & 1849	& etoinasrhlcdufmpybgwvxkjzq &     5413 \\
12 & Pierce       & 1853	& etinoarshlducfmpygwbvkxjqz &     16406 \\
13 & Buchanan     & 1857	& etionasrhlcdufpmywgvbxqjkz &     13696 \\
14 & Lincoln      & 1861,1865	& etoinasrhldcufpmywbgvkxjqz &     19340 \\
15 & Grant        & 1869,1873	& etoinarshldcufmpgywbvxkqjz &     11476 \\
16 & Hayes        & 1877	& etoinasrhlcdufpmygbwvjqxkz &     12171 \\
17 & Garfield     & 1881	& etoniasrhlducfmpgwybvkxjqz &     14477 \\
18 & Cleveland    & 1885	& etoinasrhdlcufpmygbwvxkzjq &     18480 \\
19 & Harrison     & 1889	& etoniasrhldcufpmwygbvkxjqz &     21394 \\
20 & Mckinley     & 1897,1901	& etnoiarshlducfpmygbwvxkjzq &     30179 \\
21 & T.Roosevelt  & 1905	& etoainrshldufwcgbpmvykxjzq &     4480 \\
22 & Taft         & 1909	& etoinasrhdclfumpgywbvkxjqz &     26272 \\
23 & Wilson       & 1913,1917	& etoanisrhdlucfwpmgyvbkjqxz &     14360 \\
24 & Harding      & 1921	& etnioarsldhcufmwpgybvkxzjq &     16508 \\
25 & Coolidge     & 1925	& etonairshldcufmpwybgvxkjqz &     19482 \\
26 & Hoover       & 1929	& etoinarshldcufmpgywbvzxjkq &     19256 \\
27 & F.D.Roosevelt
	& 1933,1937,1941,1945 	& etoainrshldcfumpwygvbkjxzq &     25696 \\
28 & Truman       & 1949	& etoainrshldcfumpwgyvbkjqxz &     11070 \\
29 & Eisenhower   & 1953,1957	& etoainrshldfcumpwygbvkqjxz &     18313 \\
30 & Kennedy      & 1961	& etoanrsihldfuwcmgypbvkjxzq &     6003 \\
31 & Johnson      & 1965	& etanoirshdluwcfmgybpvkjxzq &     6468 \\
32 & Nixon        & 1969,1973	& etoanirshldcuwfmpgbyvkjqxz &     17142 \\
33 & Carter       & 1977	& etaonirshldumwcfpgbyvkjqxz &     5459 \\
34 & Reagan       & 1981, 1985	& etonarishdlumwcfgpybvkjxzq &     22494 \\
35 & G.H.W.Bush   & 1989 	& etaonrishdluwcmgfybpvkzjxq &     9781 \\
36 & Clinton      & 1993,1997 	& eotanrishldcumwfpgybvkjzxq &     16915 \\
37 & G.W.Bush     & 2001,2005 	& etonairsdhlcufmwygpbvkjzqx &    16759 \\
38 & Obama        & 2009	& etoanrsihdlucwfmgypbvkjqxz &     10632 \\
\hline 
\end{tabular}
\caption{
\label{table1}
The names of the 38 U.S. presidents, the years of their inaugural 
speech, the order of letters ranked by their frequency in
the corresponding president's speech, and the total counts of letters.
}
\end{table}

In Table \ref{table1} English letters are sorted by their frequency of usage, from common
to rare, for the 38 US president speeches. Again, {\sl e} and {\sl t} are
consistently ranked as number 1 and 2 (with the exception of
Clinton's speech, where {\sl o} is ranked second), but the ranking
order of {\sl a } and {\sl i} seem to change with time:
in older speeches (e.g. before year of 1890), {\sl i} is 
ranked higher than {\sl a}, after 10 more presidents where
{\sl i} and {\sl a} were used about equally, then the order is 
reversed for newer speeches (e.g. after the year 1960).

\begin{table}[ht]
\begin{tabular}{ccccc}
\hline
 & name & years of speech & sorting of alphabets & num \\
\hline
1 & Carranza       & 1917,1918,1919               & eaosnirdlctupmbgyvfqhjxz\~{n}kw &     539107 \\
2 & De la Huerta   & 1920 	                  & eaosinrdlctupmbgfvyhqjzx\~{n}kw &     113057 \\
3 & Obreg\'{o}n    & 1921,1922,1923,1924          & eaosinrdlctupmbgyfvhqjxz\~{n}kw &     675552 \\
4 & El\'{i}as 	   & 1925,1926,1927,1928          & eaosinrdlctupmbgyfvqhjzx\~{n}kw &     700715 \\
5 & Portes Gil     & 1929                         & eaosinrdlctupmbgyvfqhjzx\~{n}kw &     231873 \\
6 & Ortiz          & 1930,1931,1932               & eaoisnrdlctupmbgvfyqhjzx\~{n}kw &     664319 \\
7 & Rodr\'{i}uez  & 1933,1934                    & eaoisnrdlctupmbgyvfqhjzx\~{n}kw &     301745 \\
8 & C\'{a}rdenas   & 1935,1936,1937,1938,1939,1940& eaosinrldctupmbgvyfqhjxz\~{n}kw &     402748 \\
9 & \'{A}vila      & 1941,1942,1943,1944,1945,1946& eaosinrlcdtumpbygvfqhjzx\~{n}kw &     734540 \\
10 & Alem\'{a}n    & 1947,1948,1949,1950,1951,1952& eaoisnrcltdumpbyvgfhqzjx\~{n}kw &     549980 \\
11 & Ruiz          & 1953,1954,1955,1956,1957,1958& eaosinrldctumpbygvfqhjzx\~{n}kw &     592550 \\
12 & L\'{o}pez     & 1959,1960,1961,1962,1963,1964& eaosinrldctupmbgvyfhqzjx\~{n}kw &     712056 \\
13 & D\'{i}az      & 1965,1966,1967,1968,1969,1970& eaosinrldctupmbgvyfqhzjx\~{n}kw &     785528 \\
14 & Echeverr\'{i}a& 1971,1972,1973,1974,1975,1976& eaosinrldctumpbvgfyqhjzx\~{n}kw &     792338 \\
15 & L\'{o}pez Portillo     
                   & 1977,1978,1979,1980,1981,1982& eaosinrlcdtumpbygvfqhzjx\~{n}kw &     684658 \\
16 & De la Madrid  & 1983,1984,1985,1986,1987,1988& eaoisnrlcdtumpbyvgfhqzjx\~{n}kw &     761274 \\
17 & Salinas       & 1989,1990,1991,1992,1993,1994& eaosinrlcdtumpbvygfhqzxj\~{n}kw &     624933 \\
18 & Zedillo       & 1995,1996,1997,1998,1999,2000& eaosinrlcdtumpbgyvfqhzjx\~{n}kw &     282463 \\
19 & Fox           & 2001,2002,2003,2004,2005     & eaosinrldctumpbgyvfqhzjx\~{n}kw &     311429 \\
\hline \\
\end{tabular}
\caption{
\label{table2}
The last names of the 19 Mexican presidents, the years when they addressed
the congress, the order of letters ranked by their frequency in
the corresponding president's address, and the total counts of letters.
}
\end{table}

\large

Table \ref{table2} shows the corresponding sorting of Spanish letters
in the 19 Mexico president addresses. The sequence 
{\sl eaosinr} consistently appears at the head of
the string. However, the order of {\sl d} and {\sl l}
has been switched from {\sl dl} in the first half of 20th century
(until president Rodr\'{i}uez whose term ended in 1934)
to {\sl ld} in the second half of the century (since
president Alem\'{a}n whose term started in 1946).

\begin{figure}[th]
\begin{center}
  \begin{turn}{-90}
   \epsfig{file=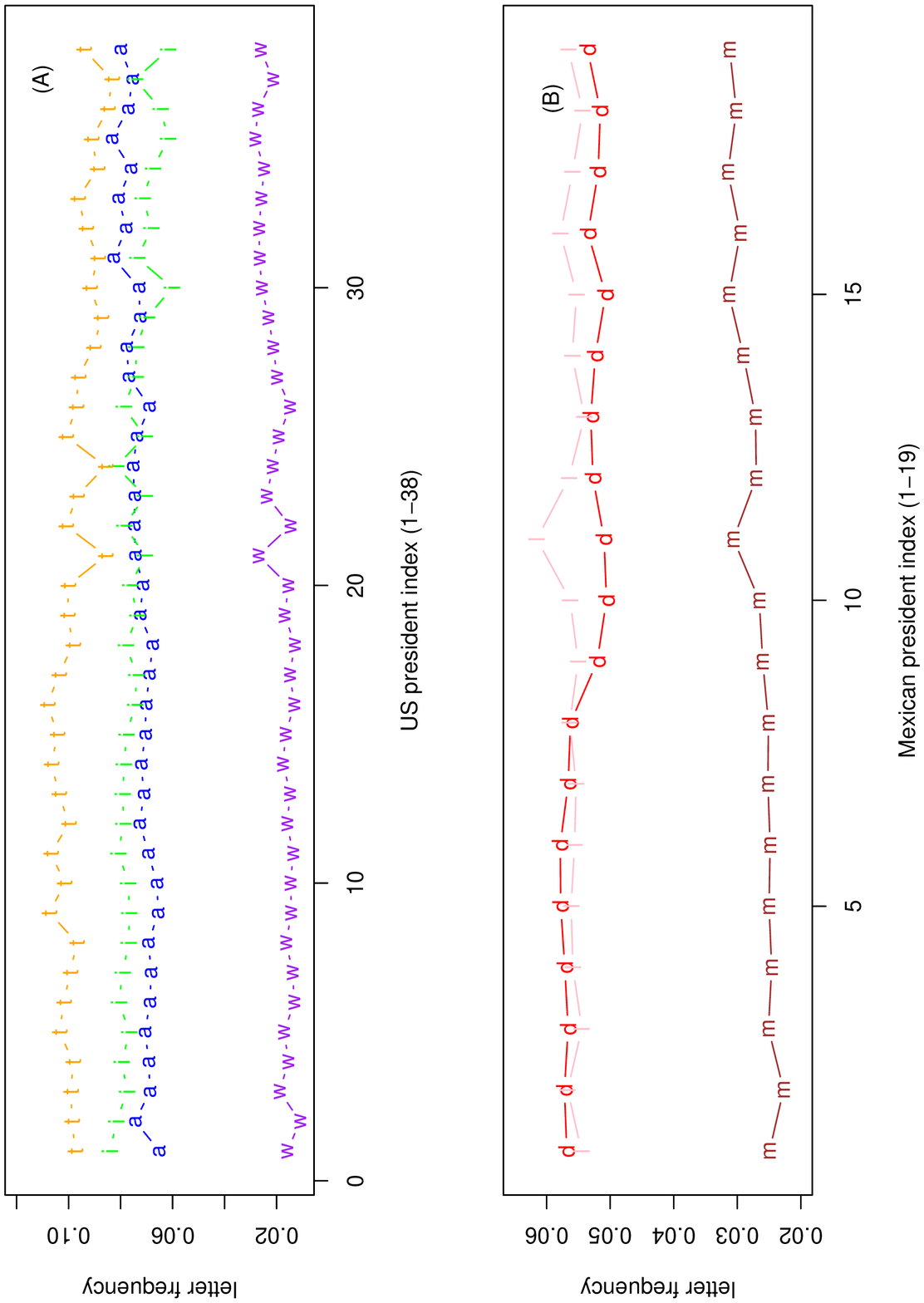, width=10cm}
  \end{turn}
\end{center}
\caption{
\label{fig2}
Temporal change of frequency in selected letters.
(A) Letters {\sl t, i, a, w} in 38 U.S. presidential speeches.
(B) Letters {\sl d,l,m} in 19 Mexican presidential speeches.
}
\end{figure}

To confirm the observation from Fig.\ref{fig1} and Tables \ref{table1},\ref{table2}, 
in Fig.\ref{fig2} we directly plot the English letter frequencies of {\sl t,w,a,i}
and Spanish letter frequencies of {\sl d,l,m}. 
Indeed, there is higher usage of {\sl w} and lesser usage
of {\sl t} in recent US president speeches, and the relative order
of {\sl a} and {\sl t} was switching from year 1889 to 1957.
For Mexican president addresses, the letter {\sl l}
overcomes {\sl d} in the last few decades. There is
also an upward trend for the usage of Spanish letter {\sl m}.

Despite these interesting trends of a few letters for the last
two hundreds of years for English and one hundred of
years in Spanish, the overall letter frequencies remain 
more or less stable. We combine all 38 English files into one
(and 19 Spanish files into one) to examine the rank frequency 
distribution.

\large

\section{Fitting ranked letter frequency distributions}

\indent

We used ten different functions to fit the ranked letter
frequency distribution in US presidential inaugural speeches that 
is averaged over all 38 presidents, and Mexican presidential
addresses to the congress averaged over 19 presidents. Here is a list of these
functions ($f$ denotes the normalized letter frequency, $r$ denotes
the rank: $r=1$ for most frequent letter and $r=26$ (or 27)
for the rarest letter, and $n=26,27$ is the maximum rank value):
\begin{eqnarray}
\label{eq-all}
\mbox{Gusein-Zade} &:& f= C  \log \frac{n+1}{r}  \\
\mbox{power-law} &:& f= \frac{C}{r^a} \\
\mbox{exponential} &:& f= C e^{-ar} \\
\mbox{logarithmic} &:& f= C - a \log(r)  \\
\mbox{Weibull} &:& f= C \left( \log \frac{n+1}{r}\right)^a \\
\mbox{quadratic logarithmic} &:& f= C - a \log(r) - b \left( \log(r) \right)^2  \\
\mbox{Yule} &:& f= C \frac{b^r}{r^a} \\
\mbox{Menzerath-Altmann/Inverse-Gamma} &:& f= C \frac{e^{-b/r}}{r^a} \\
\mbox{Cocho/Beta} &:& f= C \frac{ (n+1-r)^b}{r^a} \\
\mbox{Frappat} &:& f= C + br + c e^{-ar} 
\end{eqnarray}
Since $f$ is the normalized frequency, $\sum_{i=1}^n f_i =1$, which
adds a constrain on one parameter. The parameter under constraint
is labeled as $C$ whose value is generally of no interest to us. Besides $C$,
the number of free (adjustable) parameters in these fitting functions
ranges from 0 (Gusein-Zade) to 3 (Frappat). The power-law, exponential,
logarithmic, and Weibull  functions have 1 free parameter, 
quadratic-logarithmic, Yule, Menzerath-Altmann/Inverse-Gamma, and Cocho/Beta,
functions have 2 free parameters, as discussed in \citep{entropy10}

The power-law (Eq.(2)) and exponential function (Eq.(3)) are 
often the first group of function to be tested, due to their simplicity
and widespread applicability.
The zero-free-parameter function (Gusein-Zade) in Eq.(1) \citep{gusein1,gusein2,gusein3} 
actually
corresponds to the exponential cumulative distribution,
and the Weibull function (Eq.(5)) \citep{nabeshima} corresponds to the
stretched exponential cumulative distribution. The conversion
from cumulative distribution to rank distribution of these two
functions are discussed in details in \citep{entropy10}.

The logarithmic function (Eq.(4)) is an extension of the Gusein-Zade
function $C\log(n+1)-C\log(r)$ by allowing the coefficient of
$\log(r)$ term to be independently fitted. Then the quadratic
logarithmic function is an extension of the logarithmic function
by adding one extra term. The logarithmic function is mentioned
in \citep{kanter,vlad}, whereas quadratic logarithmic function has
not been used to the best our knowledge.

The three two-parameter functions used are all attempts to
modify the power-law function: Yule function \citep{yule25,martindale} uses an 
exponential function ($b^r$), Menzerath-Altmann or inverse-Gamma 
function \citep{altmann} uses an exponential function 
of the inverse of rank ($e^{-b/r}$), and Cocho or Beta function
\citep{beta1,beta2,beta3}
uses a power-law function of the reverse rank ($(n+1-r)^b$).
The 3-parameter function in Eq.(8) proposed in \citep{frap,frap2} is to add a linear 
trend over the exponential function.

All $x$ and $y$ relationship in Eqs.(1-10) are non-linear. It is possible
to transform variables or introduce new variables to carry
out the fitting by multiple linear regression. For example, 
after define $y'=\log(f)$, $x'_1=\log(r)$, $x'_2=\log(n+1-r)$, 
the Cocho/Beta function is equivalent to a multiple
regression $y'= c_0 +c_1 x_1 +c_2 x_2$, where
the regression coefficients can be converted back to the
parameters used in Eq.(7): $C=e^{c_1}, a=-c_1, b=c_2 $.

The data-fitting result in the transformed variable, however,
is generally not identical to the result in its original
nonlinear form. Our method is to first use the multiple linear 
regression in the transformed version, if possible, in
order to obtain a rough estimation of the parameter values. 
Then these values are used as the initial condition for 
nonlinear least-square iteration 
(using the {\sl nls} function \citep{nls} in R:
{\sl http://www.r-project.org/}).

\begin{figure}[th!]
\begin{center}
  \begin{turn}{-90}
   \epsfig{file=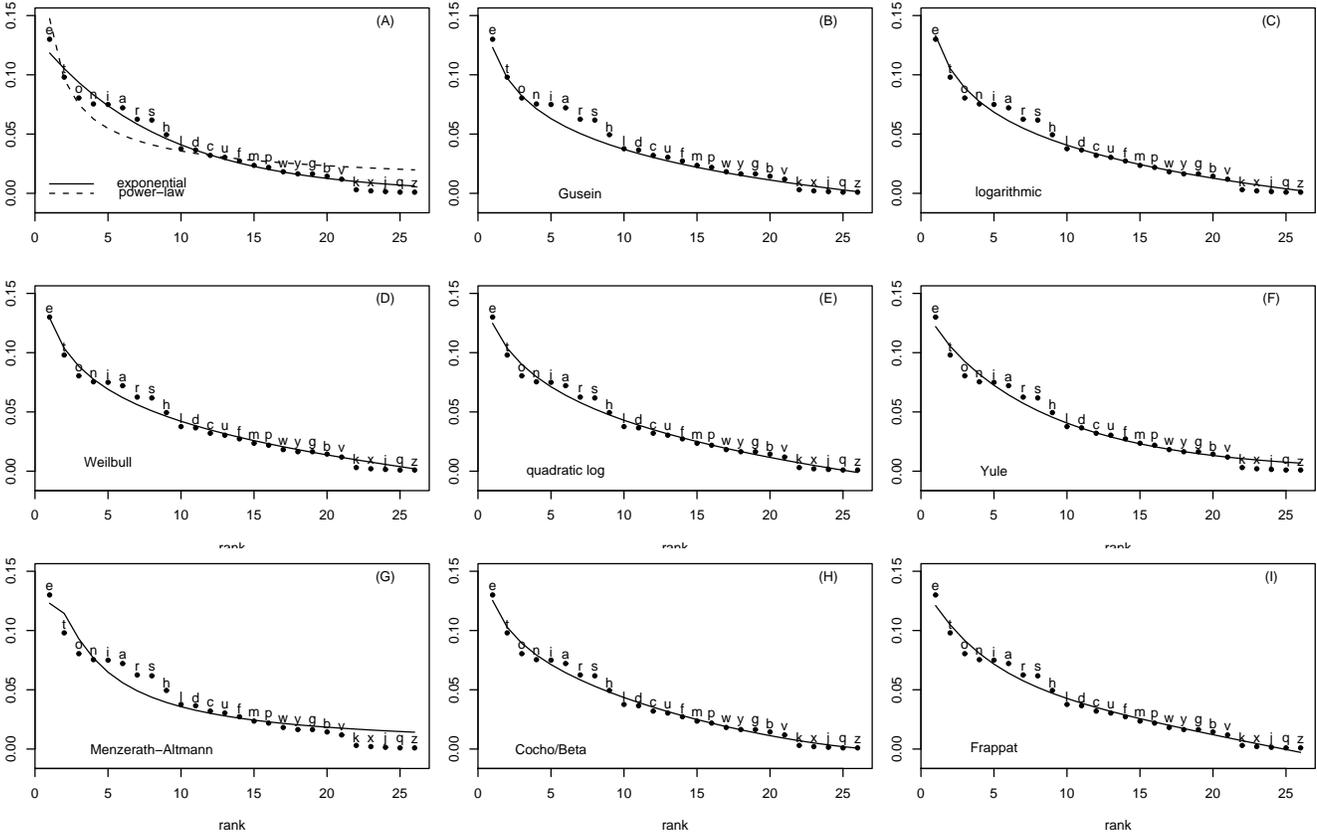, width=11cm}
  \end{turn}
\end{center}
\caption{
\label{fig3}
Fitting ranked English letter frequency of U.S. presidential
speech by ten different functions: 
(A) power-law ($a=0.616$) and exponential function ($a=0.118$);
(B) Gusein function ($C=0.0374$);
(C) logarithmic function ($a=0.0401$);
(D) Weibull function ($a=0.935$);
(E) quadratic logarithmic function ($a=0.0280$, $b=0.00325$);
(F) Yule function ($a=0.0543$, $b=0.897$);
(G) Menzerath-Altmann/Inverse-Gamma function ($a=-1.05$, $b=-1.31$);
(H) Cocho/Beta function ($a=0.210$, $b=1.35$);
(I) Frappat function ($a=0.245$, $b=-0.00242$, $c=0.0813$).
The fitting performance measured by SSE and AIC/BIC is shown
in Table \ref{table3}.
}
\end{figure}

\begin{figure}[th!]
\begin{center}
  \begin{turn}{-90}
   \epsfig{file=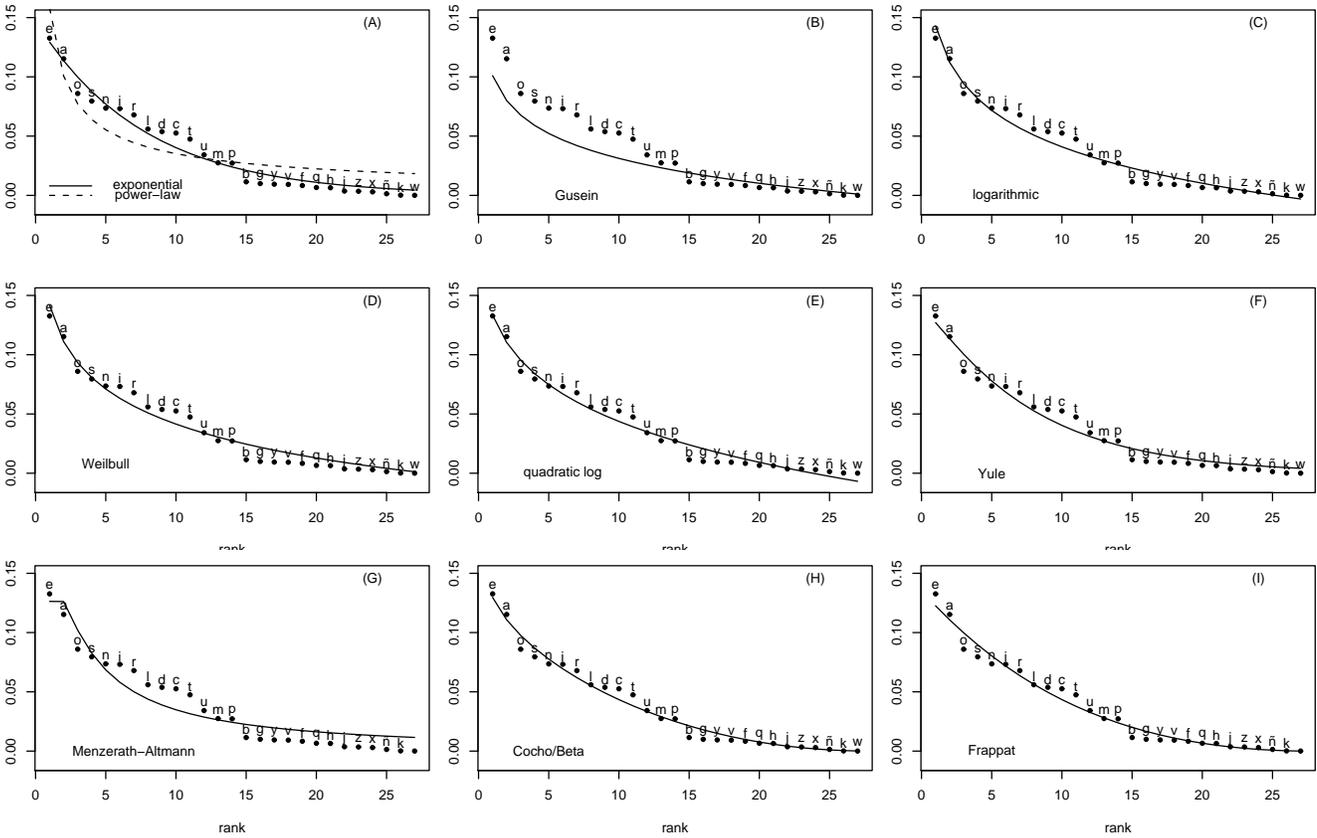, width=11cm}
  \end{turn}
\end{center}
\caption{
\label{fig4}
Fitting ranked Spanish letter frequency of Mexican presidents'
speech to congress by ten different functions:
(A) power-law ($a=0.653$) and exponential function ($a=0.130$);
(B) Gusein function ($C=0.0303$);
(C) logarithmic function ($a=0.0443$);
(D) Weibull function ($a=1.05$);
(E) quadratic logarithmic function ($a=0.0306$, $b=0.00362$);
(F) Yule function ($a=-0.0333$, $b=0.873$);
(G) Menzerath-Altmann/Inverse-Gamma function ($a=-1.22$, $b=-1.69$);
(H) Cocho/Beta function ($a=0.115$, $b=2.04$);
(I) Frappat function ($a=0.0592$, $b=0.00315$, $c=0.276$).
The fitting performance measured by SSE and AIC/BIC is shown
in Table \ref{table3}.
}
\end{figure}

\begin{figure}[th!]
\begin{center}
  \begin{turn}{-90}
   \epsfig{file=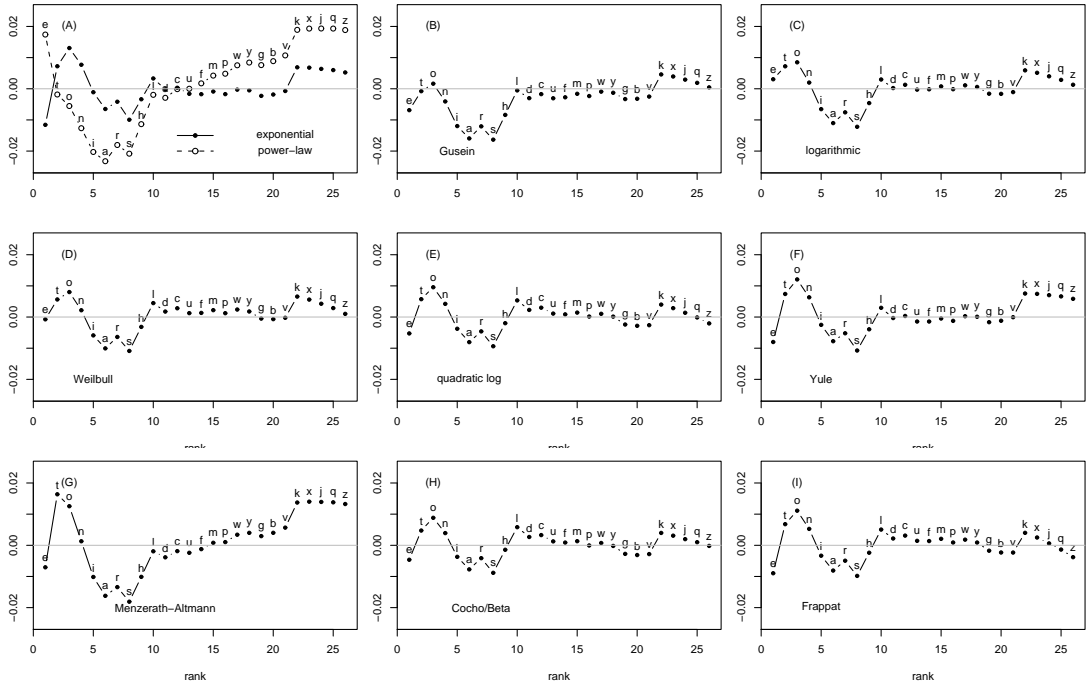, width=9cm}
  \end{turn}
\end{center}
\caption{
\label{fig5}
Fitting errors (residual, deviance), $y_{(r)}-f(r)$, of the
ten functions used in Fig.\ref{fig3} for U.S. presidential
speeches.
}
\end{figure}

\begin{figure}[th!]
\begin{center}
  \begin{turn}{-90}
   \epsfig{file=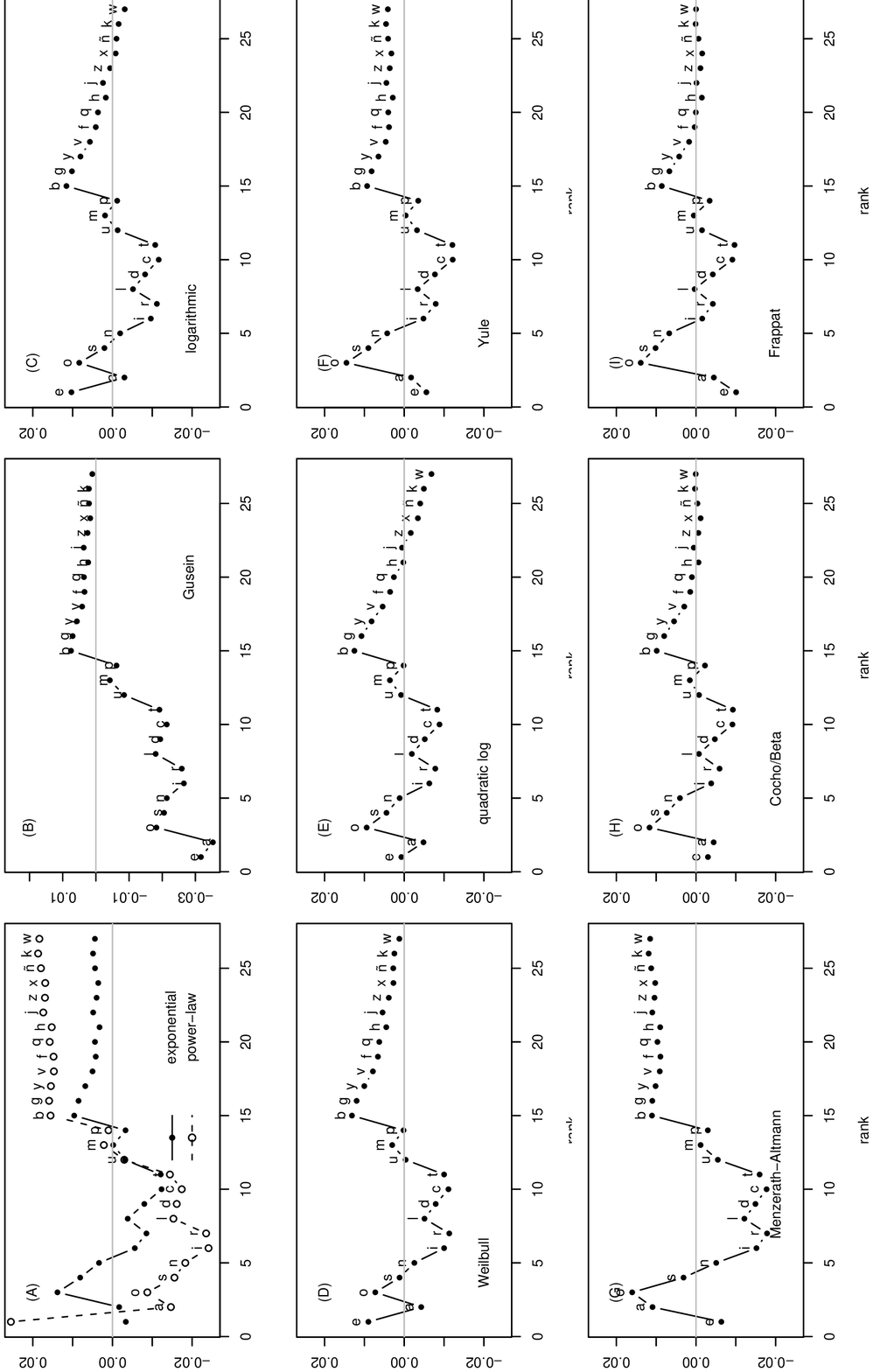, width=9cm}
  \end{turn}
\end{center}
\caption{
\label{fig6}
Fitting errors (residual, deviance), $y_{(r)}-f(r)$, of the
ten functions used in Fig.\ref{fig4} for Mexican presidents'
speech to congress.
}
\end{figure}

Fig.\ref{fig3} shows the nonlinear least-square fitting
of English letter ranked frequencies with all ten functions 
in Eqs.(1-10), and Fig.\ref{fig4} shows the result for
Spanish ranked letter frequencies. The first impression
of Figs.\ref{fig3},\ref{fig4} is that all functions seem to fit
the ranked letter frequency well, with the exception of
power-law and Menzerath-Altmann functions. Is it possible to
further distinguish those with even better fitting
performance? That is the issue to be addressed in
the next section.

\section{Comparison of the fitting performance}

\indent

How well a function $f$ fits the data can be measured by
the sum of squared errors (residuals) SSE:
\begin{equation}
SSE= \sum_{i=1}^n (y_i - \hat{f}(x_i))^2
\end{equation}
where the parameters of the function are estimated by
least-square or maximum likelihood method. It is not
correct to compare two functions with different number of
parameters, as the function with more parameters has
more freedom to adjusting in order to achieve a higher
fitting performance. In the extreme example, a function with 
unlimited number of parameters can fit a finite dataset
perfectly: this overfitting situation is called saturation.

To compare two functions with different number of
parameters, the Akaike Information Criterion (AIC)
\citep{aic} and Bayesian Information Criterion (BIC)
\citep{bic} can be used for model selection. Both criteria
discount the (log) maximum likelihood of the fitting
model by a term proportional to the number of parameters ($p$):
AIC uses the term 2$p$, and BIC uses the term $\log(n)p$
(where $n$ is the sample size). Maximizing the discounted
maximum likelihood is our criterion for the best model
(equivalent to minimizing AIC or BIC) \citep{aic-book}.

In regression models (linear or nonlinear), there is
a simple relationship between AIC/BIC and SSE if
we assume the variance of errors is unknown (and has
to be estimated from the data), and if we assume the
variance of the error is the same for all data points (details are
in Appendix).

\begin{table}[ht]
\begin{center}
\begin{tabular}{c|c|c|ccc|ccc}
\hline 
function & Eq. & $p$ & \multicolumn{3}{c|}{English} &
 \multicolumn{3}{c}{Spanish} \\
\cline{4-9}
 & & & SSE & $\Delta$ AIC & $\Delta$ BIC & SSE & $\Delta$ AIC & $\Delta$ BIC \\
\hline 
Gusein-Zade & 1 & 0 & 0.00106 & 20.2 & 17.7 & 0.00670 & 57.3 & 54.8 \\
power-law   & 2 & 1 & 0.00461 & 60.3  & 59.0 & 0.00721  & 61.3  & 60.0  \\
exponential & 3 & 1 & .000814 & 15.2  & 14.0  & 0.00118  & 12.5  & 11.2   \\
logarithmic & 4 & 1 & .000635 & 8.75 & 7.49 & 0.00115 & 11.7 & 10.4 \\
Weibull     & 5 & 2 & .000559 & 7.45 & 7.45  & 0.00136 & 18.2  & 18.2  \\
quadratic log&6 & 2 & {\bf .000460} & 2.40 & 2.40 & .000915 & 7.59 & 7.59 \\
Yule        & 7 & 2 & .000788 & 16.4 & 16.4  & 0.00117  & 14.3  & 14.3  \\
Menzerath-Altmann/Inverse-Gamma 
            & 8 & 2 &  0.00251 & 46.5 & 46.5  & 0.00340 & 43.0  & 43.0   \\ 
Cocho/Beta  & 9 & 2 & {\bf .000420}  & 0 &  0 & {\bf .000691}  & 0  & 0 \\ 
Frappat     &10 & 3 &  .000587 & 10.7 &  12.0 & .000838  & 7.20  & 8.49 \\
\hline 
\end{tabular}
\end{center}
\caption{
\label{table3}
Regression diagnosis and model selection of ten functions
on English and Spanish letter rank-frequency plots.
}
\end{table}

Table \ref{table3} shows the AIC model selection result for the fitting
in Fig.\ref{fig3} and Fig.\ref{fig4}. The best function for both English
and Spanish, selected by either AIC or BIC, is the
Cocho/Beta function (Fig.\ref{fig3}(H), Fig.\ref{fig3}(H)). The second best
function is the quadratic logarithmic function (Fig.\ref{fig3}(E), Fig.\ref{fig4}(E)).
For English text, these functions are followed by Weibull,
logarithmic, and Frappat functions. For Spanish texts,
the two best functions are followed by Frappat,
logarithmic, and exponential functions.

A single SSE value does not tell us whether there exist
systematic deviations (e.g., larger deviations at high
rank numbers). To address this question, Fig.\ref{fig5}
and Fig.\ref{fig6} show the deviation at any rank number
for all fitting functions, for English and Spanish respectively.
It is interesting that functions with better fitting performance
all have a similar pattern in rank-specific deviation.

\section{Piecewise functions}

\indent

The zero-parameter Gusein-Zade function corresponds
to a simple exponential cumulative distribution (CD)
(for more discussions, see \citep{entropy10}):
\begin{equation}
CD= 1- \frac{r}{n+1} = 1- e^{-f/C}.
\end{equation}
In other words, the proportion of values that are larger than 
$f_0$ is equal to $e^{-f_0/C}$.
Since Gusein-Zade function (Eq.(1)) can also be written as
$C= f/\log[ (n+1)/r]$, if we plot $f_i/\log( (n+1)/r_i)$ 
against $r_i$ ($i=1,2, \cdots n$), this function predicts
a plateau.

\begin{figure}[th!]
\begin{center}
  \begin{turn}{-90}
     \epsfig{file=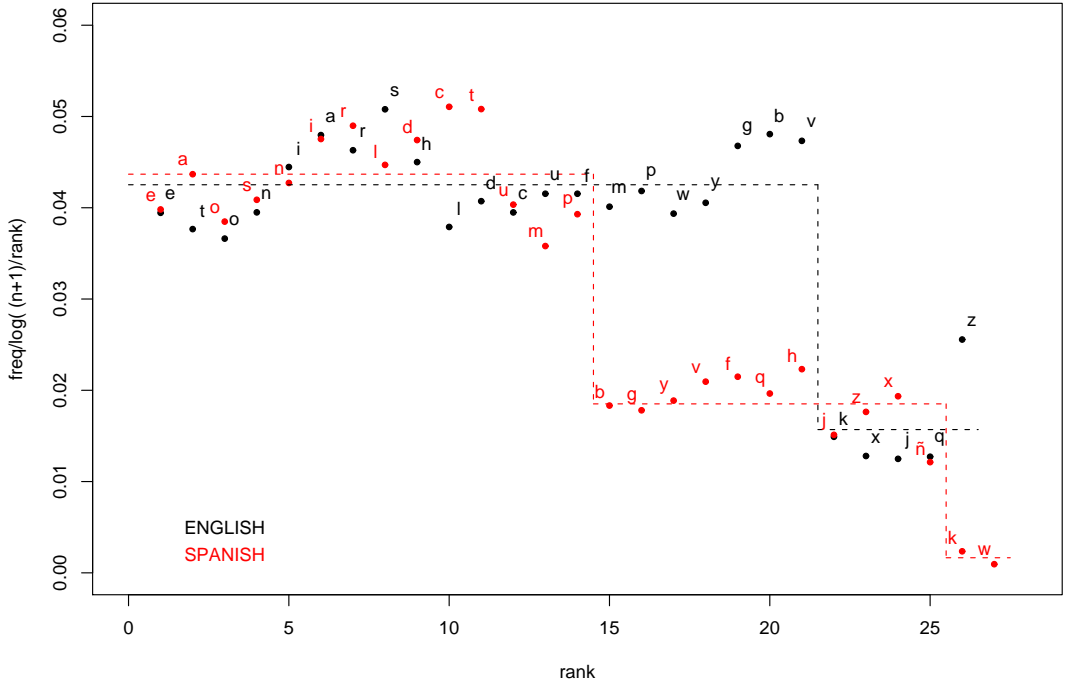, width=9cm}
  \end{turn}
\end{center}
\caption{
\label{fig7}
An alternative form of Gusein-Zade function is
$f/\log(\frac{n+1}{r})= C$, and its validity can be
checked by plotting $f/\log(\frac{n+1}{r})$ versus the
rank $r$. 
}
\end{figure}

Fig.\ref{fig7} shows $f_i/\log[ (n+1)/r_i]$ as a function
of rank, for both English (black) and Spanish (red) letters.
Surprisingly, instead of a plateau, we see step functions.
For English letters, the top 21 letters
({\sl etoniarshldcufmpwygbv}) form the first group,
and the next 5 letters ({\sl kxjqz}) form the second one. 
The average plateau height of the first group in Fig.\ref{fig7} is
0.0425, that of the second group is 0.0157.

For Spanish letters in Fig.\ref{fig7}, three groups
appear in a step function. The two rarest letters 
({\sl kw}) are very different from others (average plateau
height is 0.00165). This is a known fact as {\sl k}
and {\sl w} are only used in foreign words.  The top 14 letters ({\sl eaosnirldctump}) 
are in one group (average height of 0.0437), and the 
next 11 letters ({\sl bgyvfqhjzx\~{n}}) form the second
group (height is 0.0185). When the Spanish data is
compared to the English data, it is interesting that
the plateau height of the two groups are similar
across the language, whereas the number of letters
in the lower-plateau is much larger in Spanish than in English.

The result of Fig.\ref{fig7} indicates that we may
construct a piecewise Gusein-Zade function to fit
the ranked letter frequency distribution. It should
be noted that the number of parameters in a piecewise
Gusein-Zade function is no longer zero. For two-piece
function, three parameters are estimated:
plateau height of the first ($C_1$) and the second segment
($C_2 \ne C_1$),
and the partition position in $x$-axis ($r_0$). This minus
the normalization constraint leads to 2 free parameters.
This 3-parameter (2 of them are free) piecewise function 
can be written as:
\begin{equation}
f = \left\{
\begin{array}{ll}
C_1 log \frac{n+1}{r} & \mbox{if $r < r_0$} \\
C_2 log \frac{n+1}{r} & \mbox{if $r \ge r_0$ } \\
\end{array}
\right.
\end{equation}
For the English letter data in Fig.\ref{fig3},
$r_0$ is chosen at 22, least square regression leads
to $C_1=0.04065$ and $C_2=0.01394$, and SSE= 0.000578.
For Spanish letters, with 2-segment function partition
at $r_0=15$, $C_1=0.0424$ and $C_2=0.01897$, and 
SSE= 0.000539. Using 3-segment function, SSE is improved 
only slightly to 0.000537. These results are comparable
to the best SSE results obtained by the Beta function
(Table \ref{table3}).

\section{Discussion}

\indent

So far we have not considered space as a ``letter". The number 
of space is simply equal to the number of words ($N_{space}=N_{word}$),
and the space frequency is $p_{space}=N_{space}/(N_{space}+N_{letter})$.
For the US presidential speeches, the averaged $p_{space}$ is
0.174. For Mexican presidential speeches, the averaged $p_{space}$
is 0.162. There is a mild upward trend for $p_{space}$ in US
presidential speeches, but such a temporal pattern is missing
in Mexican texts.

When the ``space" is considered as a symbol, its frequency is
higher than any other single letters. The rank-frequency plot
with space symbol can still be fit perfectly by the Cocho/Beta
function (result not shown). The Cocho/Beta is still the best
function than others. However, the fitted coefficient values
can be quite different when space-symbol is included. For example,
for English texts, $a=0.21$ and $b=1.35$ without the space,
but $a=0.50$ and $b=0.875$ with the space symbol. 

Due to the limited range of abscissa, many functions seem to fit the
ranked letter frequency distribution very well, and any subtle
change might disturb the relative performance among fitting functions.
Take the $-\log(r)$ type functions for example, we have considered
three similar functions already, Eq.(1), Eq.(4), Eq.(5), and Eq.(6).
The quadratic logarithmic function Eq.(6) clearly outperforms Eq.(1)
and Eq.(4), and competes with Cocho/Beta function to become the best 
fitting function. We notice that in Gusein-Zade's original publication
\citep{gusein2}, he proposed a function of the form 
$f= (1/r+1/(r+1) \cdots 1/n)/n$, which also appeared in \citep{gamow}
after a random division of unit length problem by John von Neumann.
That function can be approximated by a $-\log(r)$ function.

The piecewise plateaus revealed by Fig.\ref{fig7} seem
to partition alphabets into discrete groups. For English,
rare letters {\sl k,x,j,q,z} form their own group, with
frequencies much lower than expected by the $\log((n+1)/r$ function.
For Spanish, besides the well know letter group of {\sl k,w},
we found another group with letters {\sl b,g,y,v,f,q,h, j,z,x,\~{n}}.
The height of the first plateau is about twice that of the
second plateau, for both English and Spanish. One hypothesis
is that these lower-than-expected rare alphabets were originally
paired as one letter, then each ancestral letter was split into two
letters. Two such pairs can be imagined for English (discard {\sl z}),
and five pairs for Spanish (discard {\sl \~{n}}).

Of the ten functions used in this paper, some explicitly include
the number of letters, $n$, as part of the modeling, whereas
others do not. Those with $n$ include Gusein-Zade, Weibull,
and Cocho/Beta. For some linguistic data, the value of $n$
is fundamentally undecided, for example, the number of words
in a language. It is argued that word distribution should be
better modeled by ``large number of rare events" (LNRE)
\citep{baayen}. One consequence of LNRE is that the number
of words $n$ increases with the text length (followed the Heaps' law
\citep{heaps}), making the value of $n$ uncertain. Fortunately, 
in letter frequencies, the value of $n$ is independent of the text length. 

There might be deeper reasons why Cocho/Beta outperforms
nine other functions in fitting our data. It was
suggected that when a new random variable is constructed
by allowing both addition and subtraction of 
independent and identically distributed random variables,
but within certain range, the new random variable follows
the Cocho/Beta distribution \citep{beltran}. Perhaps 
Cocho/beta function is a limiting functional form for
ranked data under a very general condition.

In conclusion, we use ten functions to fit the English
and Spanish ranked letter frequency distribution obtained from
the US and Mexican presidential speeches. Cocho/Beta function
is the best fitting function among the ten, judged by sum of
errors (SSE) and Akaike information criterion (AIC). The
quadratic logarithmic function is a close second best.
We also discover a grouping of letters in both English and
Spanish. The rarer-than-expected group in English consists
of two pairs of letters whereas that in Spanish consists of
five pairs. There is a third, even-rarer-than-expected letter
group in Spanish with {\sl k,w}, consistent with the fact that
these are only used for foreign words. Besides the Cocho/Beta
and quadratic logarithmic function, it is not conclusive whether
other functions follow a universal relative fitting performance
order. Needless to say, studying letter frequencies in other languages
could potentially answer this question.

\section*{Appendix: Relationship between AIC/BIC and SSE}

\indent

Akaike information criterion is defined as: $AIC= - 2\log \hat{L} + 2p$,
where $\hat{L}$ is maximized likelihood, $p$ is the number of
parameter in the statistic model. When a dataset is fitted
by a model, if the error is normally distributed, the likelihood
of the model is ($n$ is the number of samples, $\sigma$ is the standard
deviation of the normal distribution for the error, $\{ y_i \}$
are the data points, and $\{ \hat{y_i} \}$ are the fitted value):
\begin{equation}
L = \prod_{i=1}^n \frac{e^{ -(y_i-\hat{y})^2/2\sigma^2} }{\sqrt{2\pi \sigma^2}}
=
\frac{ e^{ -\sum_i^n (y_i-\hat{y})^2)/2\sigma^2}}
{ (2\pi \sigma^2)^{n/2} }
\end{equation}
The $\sum (y_i-\hat{y_i})^2$ term can be called SSE (sum of squared errors).

If the error variance is unknown, it can be estimated from the data:
\begin{equation}
\hat{\sigma}^2=  \frac{SSE}{n}
\end{equation}
Replacing $\sigma$ by the estimated $\hat{\sigma}$, we obtained
the maximized likelihood, which after log is \citep{ripley}:
\begin{equation}
\log(\hat{L}) = C - \frac{n}{2} \log(\hat{\sigma}^2)
= C - \frac{n}{2} \log(SSE/n)
\end{equation}
then,
\begin{equation}
AIC = n \log(SSE/n) + 2 \cdot p + const.
\end{equation}
and
\begin{equation}
BIC = n \log(SSE/n) + \log(n) \cdot p + const.
\end{equation}

\section*{Acknowledgements}

We would like to thank Osman Tuna G\"{o}kg\"{o}z for introducing
us the work by Al-Kindi.  This work was partially supported by 
UNAM-PAPIIT project IN115908. The authors wish to thank the 
hospitality of the Centro de Investigaci\'{o}n en Matem\'{a}ticas 
Aplicadas, Pachuca, M\'{e}xico, where the draft was finalized.

\end{document}